\documentclass[letterpaper, 10 pt, conference]{ieeeconf}  

\IEEEoverridecommandlockouts                              

\usepackage{stfloats}
\usepackage{multirow}
\usepackage{graphics} 
\usepackage{epsfig} 
\usepackage{mathptmx} 
\usepackage{times} 
\usepackage{amsmath} 
\usepackage{amssymb}  
\usepackage{graphicx}   
\usepackage{float}
\usepackage{placeins} 
\usepackage{subcaption}
\usepackage{cite}

\usepackage{amsmath,amssymb,mathtools}
\usepackage{booktabs}
\usepackage[ruled,vlined,linesnumbered]{algorithm2e}
\usepackage{xspace}

\usepackage{xcolor}
\usepackage{xcolor}
\usepackage{algorithmic}
\newtheorem{definition}{Definition}

\newcommand{\Touch}{\mathrm{Touch}}
\newcommand{\LeftOf}{\mathrm{LeftOf}}

\newcommand{\CloseE}{\mathrm{closeTo}_\varepsilon}
\newcommand{\FarE}{\mathrm{farFrom}_\varepsilon}
\newcommand{\Ovlp}{\mathrm{ovlp}}
\newcommand{\PartOvlp}{\mathrm{partOvlp}}
\newcommand{\EnclIn}{\mathrm{enclIn}}

\newcommand{\Between}{\mathrm{Between}}

\newcommand{\ent}[1]{\texttt{#1}}       
\newcommand{\ival}[2]{[#1,#2]}          

\usepackage{xcolor}
\definecolor{deepblue}{HTML}{003399}
\usepackage[most]{tcolorbox}
\usepackage[colorlinks=true]{hyperref}
\hypersetup{
    urlcolor = deepblue,
    linkcolor = deepblue,
    citecolor = deepblue
}
\usepackage{pgfplots}
\pgfplotsset{compat=1.17} 
\allowdisplaybreaks

\title{\LARGE \bf
NL2SpaTiaL: Generating Geometric Spatio-Temporal Logic Specifications from Natural Language for Manipulation Tasks
}

\author{
Licheng Luo, Kaier Liang, Yu Xia, Mingyu Cai
\vspace{-0.6em}
\thanks{Licheng Luo, Yu Xia and Mingyu Cai are with University of California, Riverside, USA.
        {\tt\small \{lichengl, yxia072, mingyuc\}@ucr.edu}
        }%
\thanks{Kaier Liang is with Lehigh University, Bethlehem, USA.
        {\tt\small \{kal221\}@lehigh.edu}}
}

\begin{document}

\maketitle
\thispagestyle{empty}
\pagestyle{empty}

\begin{abstract}
While Temporal Logic provides a rigorous verification framework for robotics, it typically operates on trajectory-level signals and does not natively represent the object-centric geometric relations that are central to manipulation. Spatio-Temporal Logic (SpaTiaL) overcomes this by explicitly capturing geometric spatial requirements, making it a natural formalism for manipulation-task verification. Consequently, translating natural language (NL) into verifiable SpaTiaL specifications is a critical objective. Yet, existing NL-to-Logic methods treat specifications as flat sequences, entangling nested temporal scopes with spatial relations and causing performance to degrade sharply under deep nesting. We propose NL2SpaTiaL, a framework modeling specifications as Hierarchical Logical Trees (HLT). By generating formulas as structured HLTs in a single shot, our approach decouples semantic parsing from syntactic rendering, aligning with human compositional spatial reasoning. To support this, we construct, to the best of our knowledge, the first NL-to-SpaTiaL dataset with explicit hierarchical supervision via a logic-first synthesis pipeline.
Experiments with open-weight LLMs demonstrate that our HLT formulation significantly outperforms flat-generation baselines across various logical depths. These results show that explicit HLT structure is critical for scalable NL-to-SpaTiaL translation, ultimately enabling a rigorous ``generate-and-test'' paradigm for verifying candidate trajectories in language-conditioned robotics. Project website: \href{https://sites.google.com/view/nl2spatial}{\texttt{https://sites.google.com/view/nl2spatial}}
\end{abstract}

\section{Introduction}

Grounding natural language (NL) instructions into structured spatial reasoning remains a central challenge for language-conditioned robotics~\cite{Cohenijcai2024p885}. Although large language models (LLMs) exhibit impressive linguistic reasoning, they often struggle to represent the precise geometric structures and temporal dependencies required for complex manipulation~\cite{black2024pi0,huang2024rekep}. Consequently, language-based agents frequently generate actions that appear semantically plausible but violate rigid spatio-temporal constraints. This reflects a fundamental mismatch between the unconstrained nature of NL and the formal specifications required for rigorous downstream reasoning, control, and, most importantly, task verification~\cite{Cohenijcai2024p885}.
\begin{figure}[t]
    \centering
    \includegraphics[width=0.98\linewidth]{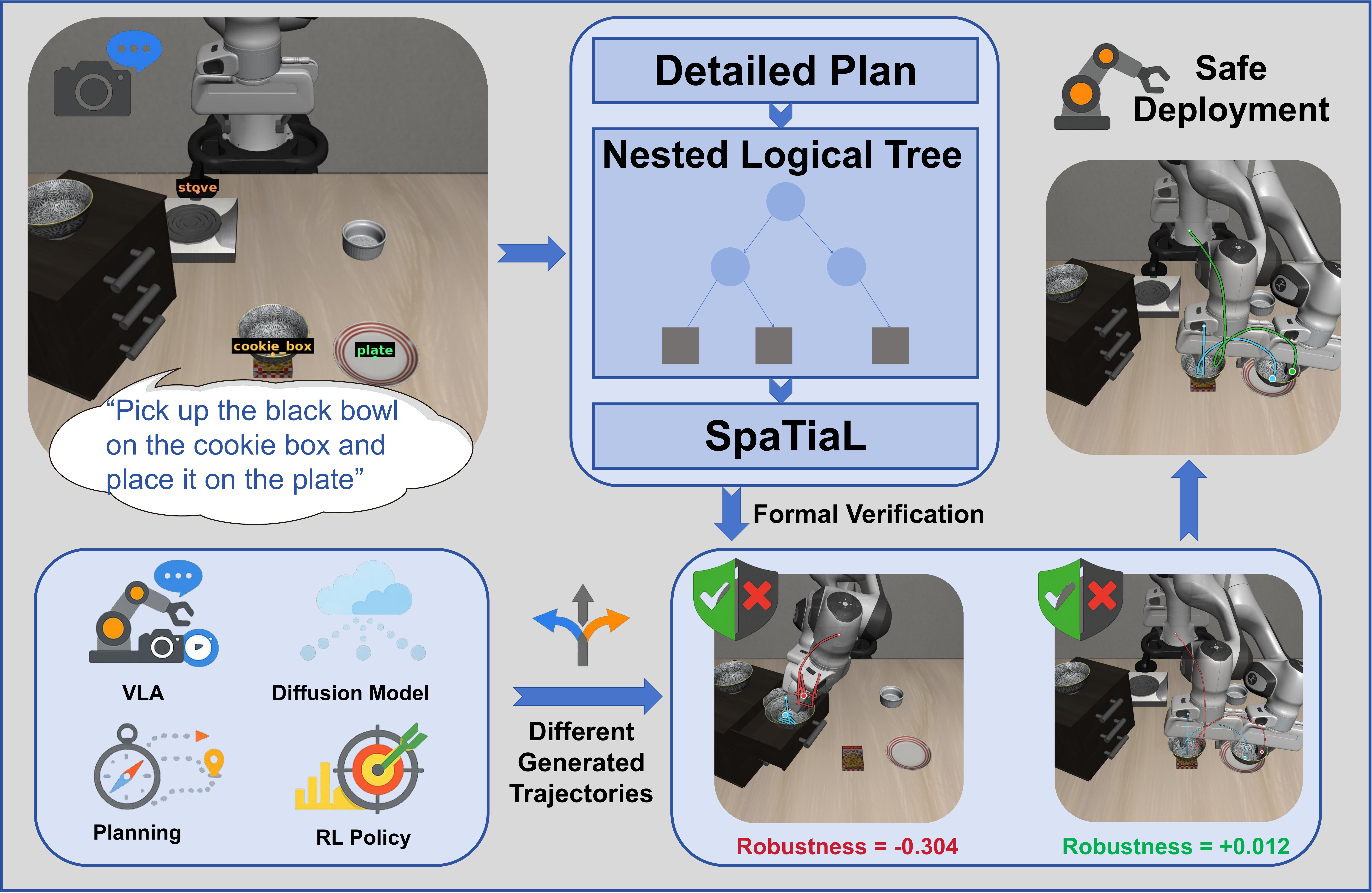}
\caption{\textbf{NL2SpaTiaL enables an interpretable generate-and-test pipeline:} The generated specifications act as executable geometric constraints to verify candidate trajectories from black-box models (e.g., VLAs). Our primary contribution is the scalable symbolic translation framework, rather than the downstream control policies.}
    \vspace{-1.5em}
\label{fig:grounding_overview}
\end{figure}
Temporal Logic (TL) has long provided a rigorous mathematical framework for verifying robotic task execution~\cite{belta2019formal,kressgazit2018synthesis}, driving recent efforts in NL-to-TL translation~\cite{brunello2019synthesis,liu2023lang2ltl,chen2023nl2tl}. However, standard TL evaluates only robot trajectories, inherently ignoring the object-level interactions crucial for manipulation. Spatio-Temporal Logic (SpaTiaL) overcomes this limitation by enabling explicit symbolic reasoning over continuous geometric relations (e.g., \texttt{Above}, \texttt{EnclIn}, \texttt{CloseE}) and their temporal evolution~\cite{pek2023spatial,spaTiaL_pkg}. Because it explicitly captures these requirements, SpaTiaL serves as a highly natural formalism for verifying complex manipulation. As illustrated in Fig.~\ref{fig:grounding_overview}, deriving executable SpaTiaL specifications from natural language unlocks an interpretable verification layer for modern generate-and-test pipelines, effectively filtering unsafe trajectories proposed by unconstrained generative models, such as vision-language-action (VLA) models~\cite{black2024pi0}.

However, reliably translating unconstrained natural language into formal SpaTiaL specifications remains a fundamental bottleneck. Existing methods typically map text to a \emph{flat} logical string~\cite{brunello2019synthesis,liu2023lang2ltl,chen2023nl2tl}. This flat formulation creates a scalability bottleneck: by coupling high-level logical routing with low-level syntactic realization, e.g., bracket matching, it makes LLMs highly prone to syntax errors and fails to preserve the natural compositional hierarchy inherent in human manipulation instructions~\cite{cosler2023nl2spec,english2025graft,mao2019nscl,ruis2020gscan}.


To address this limitation, we propose \textbf{NL2SpaTiaL}, a framework that separates semantic parsing from syntactic rendering. Rather than generating executable logic directly, the model first predicts a Hierarchical Logical Tree (HLT) as a structured JSON representation that captures the compositional structure of the instruction. A deterministic renderer then converts the validated HLT into executable SpaTiaL formulas, guaranteeing syntactic correctness by construction while preserving semantic meaning.
To mitigate data scarcity in formal-methods applications, we further develop a scalable logic-first synthesis pipeline. Instead of relying on manual annotation, the pipeline samples valid HLT structures, deterministically generates canonical instructions, and leverages LLMs for diverse paraphrasing. Although instantiated for the \textbf{NL2SpaTiaL dataset}, this abstract syntax tree (AST)–driven paradigm is logic-agnostic and readily extends to other compositional formalisms, including LTL~\cite{pnueli1977temporal}, STL~\cite{maler2004monitoring}, and task-level robotic specifications.

\noindent
\textbf{Contributions.} Our primary contributions are:

\begin{itemize}[leftmargin=*, itemsep=0pt, topsep=0pt, parsep=0pt, partopsep=0pt]
\item \textbf{A Hierarchical Translation Framework:} We introduce \textit{NL2SpaTiaL}, which bypasses the conventional flat sequence-mapping bottleneck. By predicting structured logical trees, it provides a strong inductive bias that dramatically improves reasoning robustness under deep spatio-temporal nesting.

\item \textbf{A Universal Logic-First Data Engine:} We formulate a logic-first synthesis paradigm that guarantees the structural and logical correctness of the generated supervision signal while preserving linguistic diversity. Utilizing this engine, we release the first hierarchically supervised NL-to-SpaTiaL benchmark, empirically exposing the sharp degradation of prior flat-generation methods as task complexity scales.

\item \textbf{Extreme Sample Efficiency \& Downstream Deployment:} Extensive experiments on open-weight LLMs (3.8B--14B) demonstrate that our HLT formulation achieves superior accuracy with minimal data---models trained on just 600 examples outperform flat baselines using $5\times$ more data. We validate its practical utility by deploying the generated formulas as executable safety verifiers to filter candidate trajectories in language-conditioned generative policies e.g., VLAs.
\end{itemize}

\vspace{-0.2cm}
\section{Related Work}
\label{sec:related_work}

\noindent
\textbf{Natural Language to Temporal Logic.} 
Translating natural language (NL) into temporal logic (TL) has evolved from rule-based systems~\cite{zilka2010,brunello2019synthesis} to modular pipelines with intermediate representations and decomposition strategies~\cite{liu2023lang2ltl,pan2023}. More recently, LLM-based approaches have been applied to TL translation in multiple domains, including domain-transferable translation~\cite{chen2023nl2tl}, traffic and planning applications~\cite{tr2mtl2024,ltlcodegen2025}, retrieval-augmented formulations~\cite{fang-etal-2025-enhancing}, and grammar-constrained decoding~\cite{english2025graft}. Interactive systems such as NL2Spec~\cite{cosler2023nl2spec} and SYNTHTL~\cite{menzoda2024} further improve reliability through decomposition, verification, and iterative refinement. 
Our work differs along three axes. First, we target \emph{SpaTiaL} specifications for continuous geometric manipulation constraints rather than primarily discrete TL settings. Second, we study \emph{one-shot hierarchical generation} (nested JSON HLT) rather than multi-step interactive decomposition at inference time. Third, we pair this formulation with supervised fine-tuning on a logic-first synthesized dataset with explicit subformula-level supervision, enabling controlled analysis of complexity scaling and sample efficiency.

\noindent
\textbf{Spatial Reasoning and Grounding.} 
Complementary to TL translation, spatial reasoning research studies how linguistic expressions can be grounded into symbolic or geometric structure. Prior work on Spatial Role Labeling and ISO-Space maps text to structured spatial representations~\cite{kordjamshidi2010srl,pustejovsky2017isospace}, while neuro-symbolic approaches generate executable programs and demonstrate compositional reasoning in visually grounded settings~\cite{mao2019nscl,johnson2017clevr,suhr2019nlvr2,ruis2020gscan}. Our work extends this line by coupling geometric spatial predicates with temporal operators in a unified translation target. Although our empirical focus is SpaTiaL for robotic manipulation, the HLT formulation and one-shot hierarchical generation interface are structurally compatible with other temporal-logic families (e.g., STL/LTL) when paired with an appropriate deterministic renderer.

\section{Background}
This section briefly reviews STL~\cite{maler2004monitoring} and SpaTiaL~\cite{pek2023spatial}, which provide the formal foundation for NL2SpaTiaL.

\subsection{Signal Temporal Logic}
\label{subsec:background-stl}

We consider discrete-time signals $s=[s[0],\ldots,s[\ell\!-\!1]]$ with $s[t]\in\mathbb{R}^d$. STL formulas specify temporal properties over such signals. Let $\mu := a^\top s \geq b$ denote an atomic predicate. The STL syntax is given by
\begin{equation}\label{eq:stl-syntax}
\phi \;::=\; \mu \mid \neg\phi \mid \phi_1 \wedge \phi_2 \mid \phi_1 \vee \phi_2 \mid F_{[a,b]}\phi \mid G_{[a,b]}\phi \mid \phi_1\,U_{[a,b]}\,\phi_2
\end{equation}
where $[a,b]$ specifies the temporal evaluation window. $F$, $G$, and $U$ represent the \emph{eventually}, \emph{always}, and \emph{until} operators, respectively.

The quantitative semantics (robustness) $r(s,\phi,t) \in \mathbb{R}$ assigns a signed satisfaction margin to a formula: $r(s,\phi,t)>0$ indicates satisfaction at time $t$, while $r(s,\phi,t)<0$ indicates violation. For an atomic predicate, $r(s,\mu,t)=a^\top s[t]-b$. Boolean and temporal operators are then lifted through min/max compositions, e.g.,
\begin{align}
r(s,\phi_1\wedge\phi_2,t) &= \min\!\bigl(r(s,\phi_1,t),\,r(s,\phi_2,t)\bigr), \label{eq:stl-robust-and}\\
r(s,G_{[a,b]}\phi,t) &= \min_{t'\in\,t+[a:b]} r(s,\phi,t'), \label{eq:stl-robust-G}
\end{align}
with other operators derived analogously~\cite{maler2004monitoring}.

\subsection{Spatial Temporal Logic}
\label{subsec:background-spatial}

While STL applies to generic real-valued signals, SpaTiaL~\cite{pek2023spatial} introduces geometric atomic predicates tailored to 2D/3D physical workspaces. At each time $t$, the state may include object poses and geometric attributes (e.g., centers, extents/radii, and orientations). SpaTiaL atoms evaluate continuous geometric relations such as distance, containment, and directional alignment.

To provide quantitative semantics compatible with STL monitoring, SpaTiaL defines a continuous robustness score $\rho$ for each spatial predicate. For illustration, consider the following representative predicates defined for objects $i$ and $j$:
\begin{itemize}[leftmargin=*, noitemsep, topsep=0pt]
    \item \textbf{Distance ($\CloseE$)}: Evaluates if two objects are within a threshold $\varepsilon_c$.
    $\rho(\CloseE(i,j)) = \varepsilon_c - \|p_i-p_j\|$
    
    \item \textbf{Containment ($\EnclIn$)}: Evaluates if object $i$ lies strictly inside $j$ with a margin $\delta$.
$\rho(\EnclIn(i,j)) = (r_j-\delta) - (\|p_i-p_j\|+r_i)$
    
    \item \textbf{Directional ($\LeftOf$)}: Evaluates if $i$ is to the left of $j$ with a separation margin $\kappa$.
    $\rho(\LeftOf(i,j)) = (x_j - r_j) - (x_i + r_i + \kappa)$
\end{itemize}
\begin{figure}[ht]
    \centering
    \includegraphics[width=0.98\linewidth]{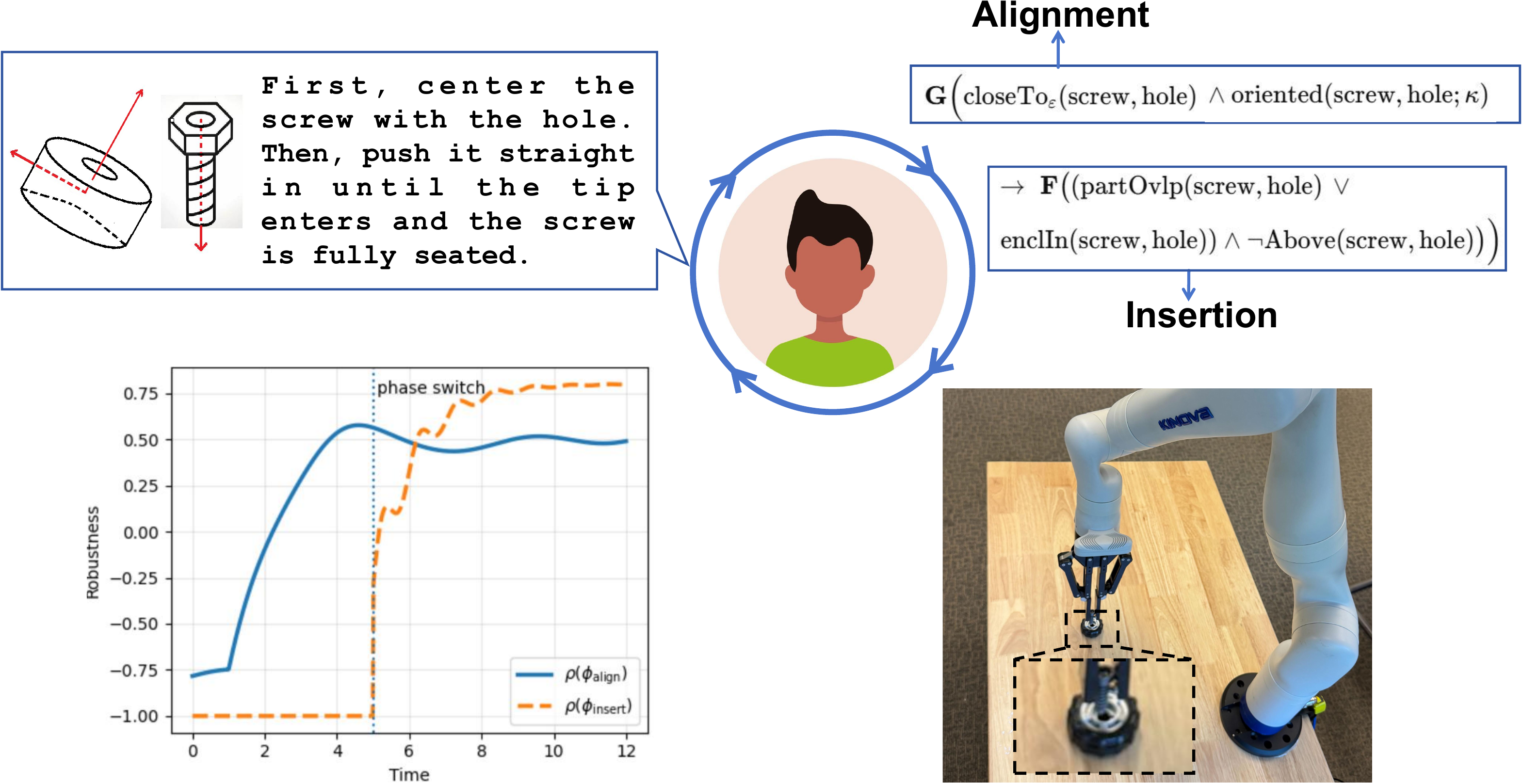}
    \caption{\textbf{SpaTiaL grounding example.} A natural-language instruction is mapped to a hierarchical SpaTiaL specification and evaluated through continuous robustness $\rho(t)$ during execution.}
    \label{fig:spatial-screw-example}
    \vspace{-1em}
\end{figure}
\begin{figure*}[t]
    \centering
    \includegraphics[width=\textwidth]{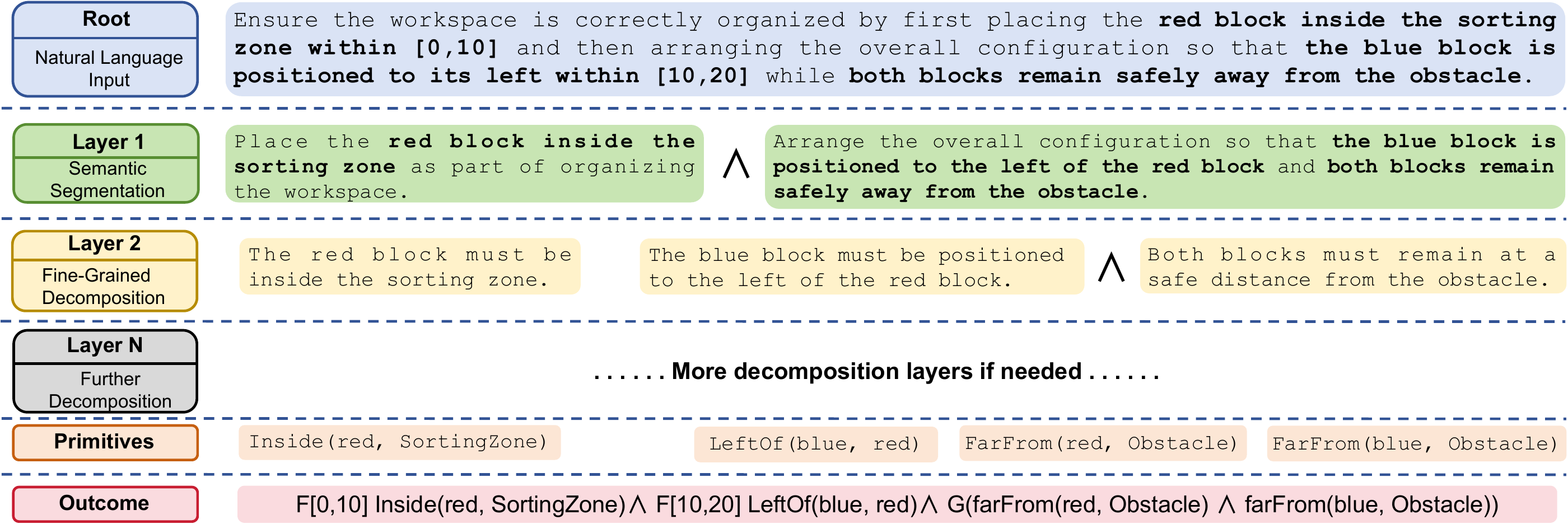}
    \caption{
    An illustrative example of the HLT representation. Instead of mapping natural language directly to a flat sequence (Outcome), NL2SpaTiaL explicitly models the semantic topology through hierarchical layers. Internal nodes dictate temporal and Boolean routing, while leaf nodes ground the geometric primitives. During inference, the LLM predicts this entire nested structure in a single shot, effectively decoupling semantic reasoning from syntactic formatting.
    }
    \label{fig:translate_framework}
    \vspace{-1em}
\end{figure*}

Other complex predicates, including boundary contact ($\Touch$), partial overlap ($\PartOvlp$), and axis-aligned betweenness ($\Between$), follow analogous geometric constructions based on signed clearances and directional projections. We refer readers to~\cite{pek2023spatial} for the complete catalog of spatial atoms and their exact quantitative semantics. Once these geometric predicates are composed through the STL operators in Eq.~\eqref{eq:stl-syntax}, they induce a trajectory-level robustness measure that evaluates not only whether a manipulation task is satisfied, but also how strongly it is satisfied or violated over time.

Formally, let $\xi = \{x_0,x_1,\ldots,x_T\}$, with $x_t \in \mathbb{R}^n$, denote a generated manipulation trajectory over a horizon $T$, where each state encodes the robot configuration together with all task-relevant objects in the workspace. Let $\mathcal{O}=\{o_1,o_2,\ldots,o_m\}$ be the set of objects involved in the task, and let the pose of object $o_i$ at time $t$ be extracted from the system state through a mapping $p_i(t)=h_i(x_t)\in SE(3)$. A manipulation task is specified by a SpaTiaL formula $\phi$, whose atomic predicates take the form $\mu_k\bigl(p_{i_1}(t),\ldots,p_{i_r}(t)\bigr): SE(3)^r \rightarrow \mathbb{R}$ and encode geometric relations among objects and the environment, such as distance constraints, containment, alignment, and collision avoidance. Evaluating the pose sequence induced by $\xi$ against $\phi$ yields the robustness value $\rho(\xi,\phi,0)$, which summarizes the satisfaction margin of the entire execution. In this way, SpaTiaL provides a unified task-level signal for continuously monitoring object--object and object--environment interactions, and, when embedded in a differentiable pipeline, for guiding optimization toward behavior that better satisfies the intended manipulation objective.

\noindent\textbf{Example.}
Fig.~\ref{fig:spatial-screw-example} illustrates how a natural-language instruction can be mapped to a hierarchical SpaTiaL specification and then evaluated through robustness during execution. The resulting predicates encode explicit geometric constraints, such as alignment, proximity, and placement, while the associated robustness signal quantifies their satisfaction margins over time. This example highlights why accurate NL-to-SpaTiaL translation is valuable for downstream monitoring, verification, and control-oriented evaluation.
 
\section{NL2SpaTiaL Framework}
\label{sec:framework}

The fundamental challenge in translating natural language instructions $x \in \mathcal{X}$ into Spatio-Temporal Logic specifications $\phi \in \Phi$ is the structural mismatch between unconstrained linguistic syntax and rigid formal logic. Standard sequence-to-sequence formulations learn a direct mapping $P(\phi \mid x)$, which couples high-level semantic composition (e.g., temporal routing and object grounding) with low-level syntactic realization (e.g., bracketing and operator formatting).

To address this mismatch, we introduce NL2SpaTiaL, which explicitly decouples semantics from syntax. We represent the target specification as a hierarchical logical tree (HLT) and use a two-stage inference process: (i) a neural semantic parsing stage that predicts the discrete tree structure, and (ii) a deterministic rendering stage that maps the validated tree to a syntactically well-formed SpaTiaL formula.

\subsection{HLT Representation}
\label{subsec:hierarchy}

SpaTiaL specifications are compositional: quantitative spatial predicates are nested under Boolean and temporal operators. To preserve this topology explicitly during generation, we represent the target as a hierarchical logical tree (HLT).

\begin{definition}[Hierarchical Logical Tree]
Let $O_{\mathrm{temp}}=\{G,F,U\}$ be temporal operators, $O_{\mathrm{bool}}=\{\wedge,\vee,\neg,\rightarrow\}$ be Boolean operators, and $P$ be the set of atomic spatial predicates. An HLT is a directed rooted tree
$T=(V_{\mathrm{int}},V_{\mathrm{leaf}},E,op,intv,args)$,
where $V_{\mathrm{int}}$ and $V_{\mathrm{leaf}}$ are disjoint sets of internal and leaf nodes, and
$E \subset V_{\mathrm{int}} \times (V_{\mathrm{int}} \cup V_{\mathrm{leaf}})$
defines parent-child scope relations.

Each node has a label $op(\cdot)$: for $v \in V_{\mathrm{int}}$, $op(v)\in O_{\mathrm{temp}}\cup O_{\mathrm{bool}}$; for $v \in V_{\mathrm{leaf}}$, $op(v)\in P$.
Temporal nodes may additionally carry an interval $intv(v)=[a,b]$ with $a,b \in \mathbb{R}_{\ge 0}$.
Each leaf node carries predicate arguments $args(v)\in \Omega^k$, where $\Omega$ denotes the universe of physical entities.
\end{definition}

This decomposition separates global logical routing (internal nodes) from local geometric grounding (leaf nodes). As a result, the HLT introduces a structural inductive bias that better matches the compositional organization of manipulation instructions than flat sequence targets.

\subsection{Inference Pipeline: Decoupling Semantics from Syntax}
\label{subsec:inference}

In flat generation, the model autoregressively emits the full specification as a 1D token sequence. As the depth of $T$ increases, errors in scope management and operator composition become more likely, since semantic decisions and syntax formatting are interleaved in the same decoding stream. We therefore decouple generation into a semantic parsing stage and a deterministic syntax projection stage.

\textbf{1. Probabilistic Semantic Parsing (Tree Generation):}
Instead of directly predicting the flat string $\phi$, we train a Large Language Model parameterized by $\theta$ as a structural parser that outputs a nested JSON representation $f_\theta:\mathcal{X}\to T_{\mathrm{JSON}}$. By constraining the target to a nested data structure, generation follows a top-down compositional process: the model first emits an internal operator node (and its interval when applicable), then recursively expands its children. This representation reduces ambiguity between temporal metadata (e.g., intervals) and leaf-level predicate arguments, and makes logical scope explicit in the output format.

\textbf{2. Deterministic Recursive Rendering:}
After $T_{\mathrm{JSON}}$ is generated, a lightweight structural validator checks format and logical consistency (e.g., operator arity, required intervals for temporal operators, and interval validity $a \le b$). The validated tree is then mapped to a SpaTiaL string by a deterministic projection $\Pi:T\to\Phi$ via recursive traversal:
\begin{equation}
\Pi(v)=
\begin{cases}
op(v)\big(args(v)\big), & v\in V_{\mathrm{leaf}},\\
op(v)\big(\Pi(v_c)\big), & op(v)=\neg,\\
op(v)_{intv(v)}\big(\Pi(v_c)\big), & op(v)\in\{G,F\},\\
\Pi(v_{c1}) \; op(v)_{intv(v)} \; \Pi(v_{c2}), & op(v)=U,\\
\left(\Pi(v_{c1}) \; op(v) \; \Pi(v_{c2}) \dots \right), & op(v)\in\{\wedge,\vee,\rightarrow\}.
\end{cases}
\end{equation}
Here, $v_c$ denotes the (ordered) child of a unary operator, and $v_{c1},v_{c2}$ denote the ordered children of a binary operator.
This decoupled architecture isolates responsibilities: the neural model $f_\theta$ resolves linguistic ambiguity and predicts structure, while the deterministic projection $\Pi$ renders syntax. Consequently, any HLT that passes structural validation yields a syntactically well-formed SpaTiaL formula by construction.

\begin{tcolorbox}[colback=blue!3!white,colframe=blue!40!black,arc=2pt,boxrule=0.5pt,
left=4pt,right=4pt,top=2pt,bottom=2pt]
\scriptsize
\textbf{Example (NL $\rightarrow$ Nested Tree $\rightarrow$ Formula).}\\
\textbf{NL:} \textit{``Between time 0 and 10, the red block is always left of the blue block.''}\\
\textbf{Nested Tree:}\\
\texttt{\{"op":"G","interval":[0,10],"children":[}\\
\texttt{\ \ \{"predicate":"LeftOf","args":["obj\_r","obj\_b"]\}}\texttt{]\}}\\
\textbf{Formula:}$\phi = G_{[0,10]}\bigl(\mathrm{LeftOf}(\mathrm{obj\_r}, \mathrm{obj\_b})\bigr)$
\end{tcolorbox}

To align base LLMs (e.g., Llama-3-8B) with our hierarchical generation paradigm, we formulate training as supervised fine-tuning (SFT) on serialized nested-tree targets. During dataset construction, we store a rich \emph{layer-based HLT} representation that includes auxiliary canonical texts and subformula metadata for synthesis and analysis. For model training, this representation is deterministically converted into a compact \emph{nested JSON tree} that retains only the logical topology (operators, intervals, predicates, and arguments) required for exact structural generation.

During training, the model is optimized with standard autoregressive next-token prediction to map a natural-language instruction to the corresponding nested JSON string. A system prompt $\mathcal{P}_{sys}$ specifies the output schema and operator constraints, thereby restricting the model to a structured SpaTiaL-compatible output space. An example prompt template and target output are shown below.

\begin{tcolorbox}[colback=green!5!white,colframe=green!40!black,arc=2pt,boxrule=0.5pt,left=4pt,right=4pt,top=2pt,bottom=2pt]
\scriptsize
\textbf{System prompt:}\\
You are a semantic parser for Spatio-Temporal Logic. Output a HLT as a single nested JSON.\\
\textbf{Format:} \textit{Internal node:} \texttt{\{"op": "<op>", "interval": [a,b], "children": [...]\}}; \textit{Leaf node:} \texttt{\{"predicate": "<name>", "args": [...] \}}.\\
\textbf{Constraints:} Temporal operators require explicit intervals. Arity bounds: G/F/Not(=1 child), U/Implies(=2), And/Or($\ge$2). Output ONLY valid JSON.
\end{tcolorbox}
\vspace{-0.5em}
\begin{tcolorbox}[colback=yellow!5!white,colframe=yellow!40!black,arc=2pt,boxrule=0.5pt,left=4pt,right=4pt,top=2pt,bottom=2pt]
\scriptsize
\textbf{User prompt:}\\
Ensure the red block stays above the blue block and touches the green block throughout time steps 5 to 20.\\
\textbf{Assistant (Target $\mathcal{T}_{\mathrm{JSON}}$):}\\
\texttt{\{"op": "G", "interval": [5, 20], "children": [}\\
\texttt{\ \ \{"op": "And", "children": [}\\
\texttt{\ \ \ \ \{"predicate": "Above", "args": ["obj\_r", "obj\_b"]\}, \{"predicate": "Touch", "args": ["obj\_r", "obj\_g"]\}}\\
\texttt{\ \ ]\} ]\}}
\end{tcolorbox}

By applying SFT directly to the structured JSON target, the training objective naturally penalizes topological violations (e.g., closing a dictionary prematurely or misplacing the \texttt{"children"} key). During inference, the generated JSON string is parsed, validated, and directly mapped to the final flat formula via deterministic rules (\textit{i.e.,} $\mathcal{T}_{\mathrm{JSON}} \rightarrow \Phi$) without requiring step-by-step intermediate translations. To systematically construct the hierarchical data pairs required for this training, we detail our logic-first synthesis pipeline in the following section.

\section{Hierarchical Data Pair Generation}
\label{sec:datasyn}

\begin{figure}[htpb]
    \centering
    \includegraphics[width=0.98\linewidth]{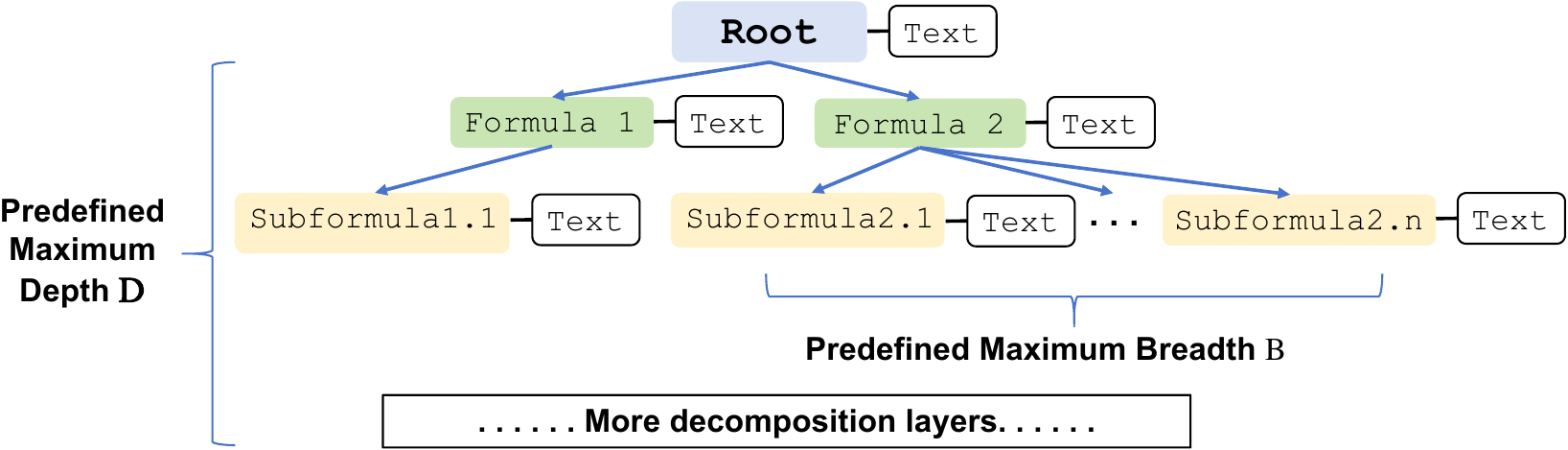}
    \caption{
    Hierarchical formula generation pipeline. We first specify structural complexity bounds (maximum depth $D$ and maximum breadth $B$), then sample concrete logical tree structures within these bounds. Leaf nodes are subsequently instantiated with lifted spatial predicates, and the resulting formula tree is back-translated in a deterministic manner, so that each logical node is paired with a corresponding canonical text description.
    }
    \label{fig:generate_tree}
    \vspace{-1em}
\end{figure}

To systematically train and evaluate NL2SpaTiaL under controlled logical complexity, we construct the dataset using a logic-first synthesis pipeline. Each example is created by first sampling a hierarchical SpaTiaL formula, then deterministically converting the formula tree into canonical natural language, and finally generating linguistically diverse paraphrases conditioned on the underlying logic. The pipeline overview is shown in Fig.~\ref{fig:generate_tree}.

\textbf{Sampling hierarchical logical skeletons.}
We first sample a rooted logical tree structure without instantiated spatial predicates. For each example, we choose a maximum depth \(D\) and assign each internal node a branching factor from a prescribed range, which determines the tree topology and the number of leaves \(L\). Internal nodes are then labeled with operators from \(\mathcal{O} = \{\neg, \wedge, \vee, \mathbf{G}_{[a,b]}, \mathbf{F}_{[a,b]}, \mathbf{U}_{[a,b]}\}\), with temporal intervals \([a,b]\) sampled from predefined ranges subject to \(0 \le a \le b\). This yields a hierarchical logical skeleton whose topology and operator types are fixed while leaves remain uninstantiated.

\textbf{Instantiating lifted spatial atoms.}
Given a skeleton with \(L\) leaves, we instantiate each leaf with a lifted spatial predicate sampled from geometry-aware templates (e.g., \(\mathrm{LeftOf}(\mathrm{obj}_i,\mathrm{obj}_j)\), \(\mathrm{EnclIn}(\mathrm{obj}_i,\mathrm{reg}_k)\)). We maintain a symbolic universe \(\mathcal{U} = \{\mathrm{obj}_1,\dots,\mathrm{obj}_M, \mathrm{reg}_1,\dots,\mathrm{reg}_K\}\) of object and region identifiers, and sample predicate arguments (and any tolerance parameters) from predefined sets or ranges. Because all identifiers are symbolic, the resulting formulas are scene-agnostic yet structurally interpretable. Substituting these atoms into the skeleton yields a complete hierarchical SpaTiaL formula \(\phi\).

\textbf{Deterministic hierarchical back-translation.}
We then convert the formula tree into canonical natural-language descriptions using a deterministic realization map \(\tau(\cdot)\) applied recursively from leaves to root. Spatial atoms are rendered with fixed templates, temporal operators introduce explicit time scopes, and Boolean operators determine clause composition. For every node \(\phi_i\) in the tree, this process produces a canonical description \(\tau(\phi_i)\). Because the mapping is grammar-driven and deterministic, each formula tree has a unique canonical realization, which avoids ambiguity during supervision generation.

\textbf{Linguistic diversification.}
To reduce template bias and improve robustness to natural-language variation, we generate paraphrased variants of each canonical description. For a node \(\phi_i\) with canonical text \(\tau(\phi_i)\), we sample variants \(\nu \sim p_{\mathrm{LLM}}(\cdot \mid \tau(\phi_i), \phi_i)\), where the exact formula \(\phi_i\) is included as conditioning context to preserve semantic fidelity. This yields stylistically diverse yet logic-consistent descriptions for both full formulas and subformulas.

\textbf{Dataset structure.}
For each sampled formula tree, the dataset stores a rich \emph{layer-based HLT} representation, including the root formula \(\phi\), its canonical description \(\tau(\phi)\), paraphrases \(\{\nu_k(\phi)\}\), and analogous tuples for internal nodes and leaves: \(\{(\phi_i, \tau(\phi_i), \{\nu_k(\phi_i)\})\}_i\). This yields dense supervision at both the global instruction level and the subformula level. For model training, the layer-based representation is deterministically converted to nested JSON targets, while the richer layer-based form remains useful for synthesis, analysis, and future multi-level learning studies.

\section{Experiments}
\label{sec:experiments}

The primary objective of our experiments is to test whether explicit structural decomposition improves NL-to-SpaTiaL translation under increasing logical complexity. We compare our HLT formulation against a flat generation baseline across four axes: model scale, linguistic diversity, logical complexity, and training data composition. Our central hypothesis is that flat generation becomes increasingly fragile under deep nesting because semantic composition and syntax rendering are entangled, whereas HLT alleviates this bottleneck through an explicit hierarchical target structure.

\subsection{Experimental Setup}

\textbf{Dataset Complexity Scaling.} 
We evaluate on benchmark sets generated by the logic-first synthesis pipeline (Sec.~\ref{sec:datasyn}), parameterized by maximum logical depth ($D$) and breadth ($B$), ranging from \texttt{D2\_B4} (shallow constraints) to \texttt{D4\_B3} (deeply nested multi-stage goals). Each evaluation set contains 150 instances constructed from 30 unique logical skeletons with 5 paraphrases per skeleton, enabling controlled measurement of structural and linguistic generalization.

\textbf{Linguistic Diversity.} 
To study robustness to language variation, we define an augmentation factor corresponding to the number of paraphrases per unique logical formula ($Aug \in \{1, 3, 5, 15\}$). We also include a ``Canonical'' condition using deterministic templates only, which helps separate gains from structural learning versus gains from linguistic diversity.

\textbf{Model Scaling \& Rigor.} 
To assess whether model capacity can substitute for structural guidance, we evaluate representative open-weight models from 3.8B to 14B parameters (Phi-3.5-mini\cite{abdin2024phi3}, Llama-3-8B\cite{llama3herd2024}, and Qwen-2.5-14B\cite{yang2024qwen25}). Fine-tuned models use an expanded context window (2560 tokens) to reduce the risk of truncating deeply nested HLT outputs. We report Correct Match (CM) as the primary metric. For zero-shot evaluation, we use greedy decoding (temperature = 0.0) for deterministic output generation. All supervised results are averaged over 3 independent runs with different random seeds $\{0,42,123\}$ to account for training variance. All supervised fine-tuning uses 4-bit quantized LoRA adapters (QLoRA\cite{dettmers2023qlora}), freezing the base model weights and updating only low-rank adapter parameters for memory-efficient training.

\subsection{Few-Shot Capability Analysis}

Before analyzing fine-tuned performance, we establish a frontier capability reference using GPT-5.2 in a few-shot ($k=5$) setting to characterize the intrinsic difficulty of NL-to-SpaTiaL translation. 

As shown in Fig.~\ref{fig:gpt_fewshot}, flat sequence generation struggles severely with deeply nested formal syntax, with GPT-5.2 accuracy dropping from 73.9\% on \texttt{D2\_B4} to a mere 18.0\% on \texttt{D4\_B3}. In contrast, the HLT scaffold acts as a powerful structural prior, yielding a +24.0\% absolute gain on the most complex benchmark. This suggests that while frontier models possess spatial reasoning capacity, the ``flat'' output format remains a primary bottleneck.

\begin{figure}[htpb]
    \centering
    \begin{tikzpicture}
        \begin{axis}[
            width=\columnwidth, 
            height=4.5cm, 
            ybar=1pt, bar width=14pt,
            ymin=0, ymax=115,
            ylabel={\textbf{Accuracy (\%)}},
            symbolic x coords={D2_B4, D3_B3, D4_B3},
            xtick=data,
            xticklabels={\texttt{D2\_B4}, \texttt{D3\_B3}, \texttt{D4\_B3}},
            x tick label style={font=\small},
            ytick={0,20,40,60,80,100},
            ymajorgrids=true, grid style={dashed, gray!40},
            enlarge x limits=0.2,
            legend style={
                at={(0.98, 0.95)}, 
                anchor=north east, 
                legend columns=-1, 
                draw=gray!70,         
                fill=white,         
                font=\small
            }
        ]
        
        \addplot[fill=gray!40, draw=gray!80] coordinates {(D2_B4, 73.9) (D3_B3, 48.0) (D4_B3, 18.0)};
        \addplot[fill=teal!70, draw=teal] coordinates {(D2_B4, 96.6) (D3_B3, 74.5) (D4_B3, 42.0)};

        \node[anchor=south, font=\scriptsize, color=gray!80!black, xshift=-7.5pt] at (axis cs:D2_B4, 73.9) {73.9};
        \node[anchor=south, font=\scriptsize\bfseries, color=teal!90!black, xshift=7.5pt] at (axis cs:D2_B4, 96.6) {96.6};

        \node[anchor=south, font=\scriptsize, color=gray!80!black, xshift=-7.5pt] at (axis cs:D3_B3, 48.0) {48.0};
        \node[anchor=south, font=\scriptsize\bfseries, color=teal!90!black, xshift=7.5pt] at (axis cs:D3_B3, 74.5) {74.5};

        \node[anchor=south, font=\scriptsize, color=gray!80!black, xshift=-7.5pt] at (axis cs:D4_B3, 18.0) {18.0};
        \node[anchor=south, font=\scriptsize\bfseries, color=teal!90!black, xshift=7.5pt] at (axis cs:D4_B3, 42.0) {42.0};

        \legend{Flat, HLT}
        \end{axis}
    \end{tikzpicture}
    \vspace{-0.5em}
    \caption{Comparison of Flat vs. HLT generation for GPT-5.2. HLT significantly mitigates performance decay as logical nesting depth increases.}
    \label{fig:gpt_fewshot}
    \vspace{-15pt}
\end{figure}

\begin{figure*}[t]
    \centering
    \includegraphics[width=\textwidth]{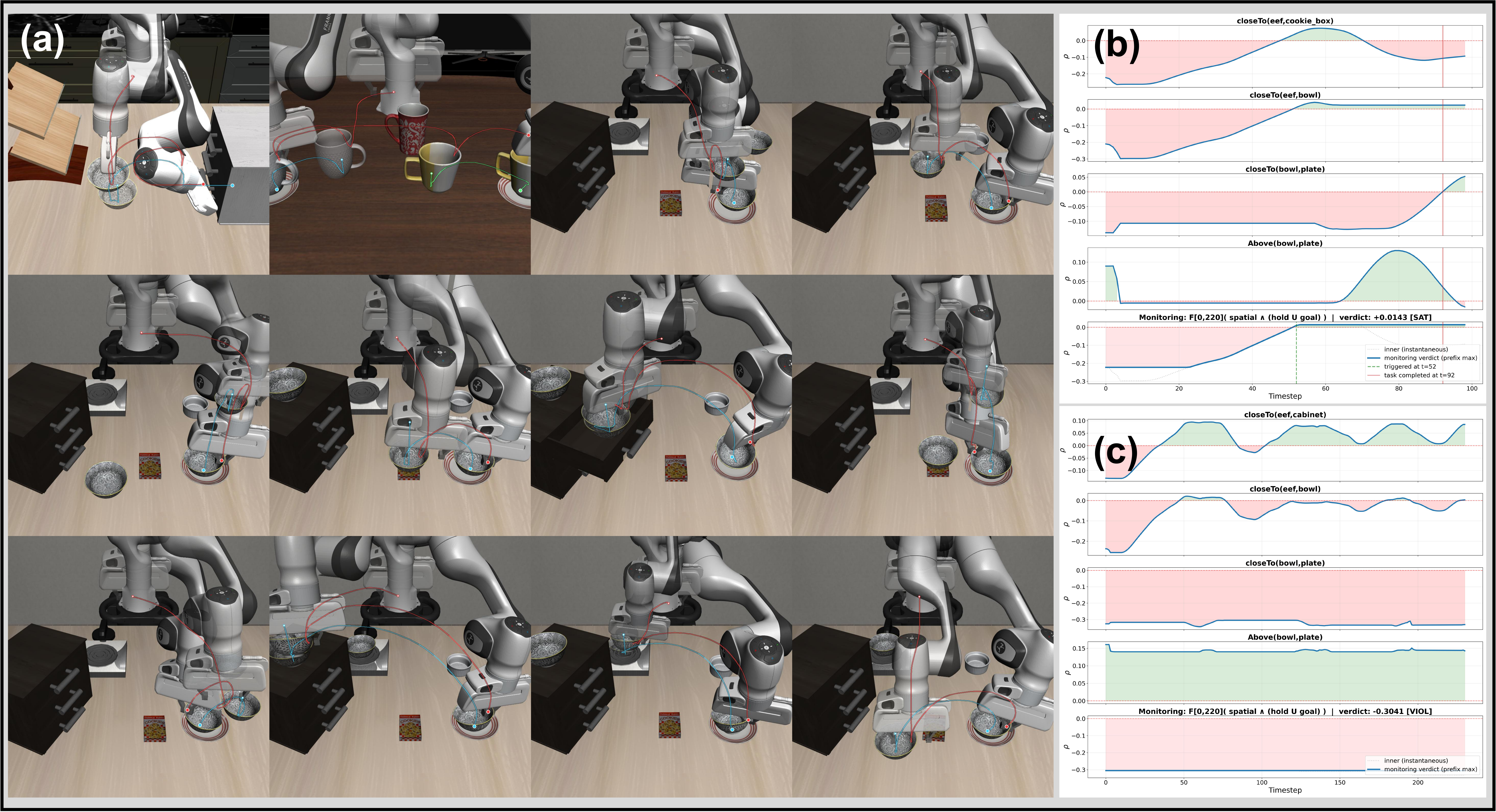}
    \caption{Evaluation of continuous SpaTiaL robustness $\rho(t)$ on the LIBERO manipulation benchmark using the $\pi_{0.5}$ Vision-Language-Action (VLA) policy. (a) Visual sequence of the multi-stage task execution. (b) Robustness analysis of a successful trajectory, demonstrating that all nested geometric and temporal constraints are rigorously satisfied, with the satisfaction margin $\rho(t) > 0$ during the specified temporal intervals. (c) The fluctuating robustness signal explicitly captures the end-effector's repeated, unsuccessful attempts to approach and grasp the bowl.}
    \vspace{-5pt}
    \label{fig:stl_evaluation_all}
\end{figure*}
\subsection{Main Results: Model Scaling vs. Structural Decomposition}

To test whether parameter scale can overcome the translation bottleneck, we fix the training distribution to the \texttt{Aug=5} setting and compare fine-tuned models from 3.8B to 14B parameters. 
\begin{table*}[t]
\centering
\caption{Model Scaling: Fine-tuned Translation Accuracy (Aug=5)}
\label{tab:model_scaling}
\resizebox{\textwidth}{!}{%
\begin{tabular}{l ccc ccc ccc}
\toprule
\multirow{2}{*}{\textbf{Model (Params)}} & \multicolumn{3}{c}{\textbf{\texttt{D2\_B4} (Shallow)}} & \multicolumn{3}{c}{\textbf{\texttt{D3\_B3} (Intermediate)}} & \multicolumn{3}{c}{\textbf{\texttt{D4\_B3} (Deeply Nested)}} \\
\cmidrule(lr){2-4} \cmidrule(lr){5-7} \cmidrule(lr){8-10}
& \textbf{Flat (\%)} & \textbf{HLT (\%)} & \textbf{$\Delta$} & \textbf{Flat (\%)} & \textbf{HLT (\%)} & \textbf{$\Delta$} & \textbf{Flat (\%)} & \textbf{HLT (\%)} & \textbf{$\Delta$} \\
\midrule
Phi-3.5-mini (3.8B) & 94.2\,{\tiny $\pm$ 0.8} & \textbf{95.8}\,{\tiny $\pm$ 0.4} & \textcolor{teal}{+1.6} & 63.1\,{\tiny $\pm$ 5.5} & \textbf{77.3}\,{\tiny $\pm$ 3.2} & \textcolor{teal}{+14.2} & 38.5\,{\tiny $\pm$ 2.4} & \textbf{56.1}\,{\tiny $\pm$ 1.8} & \textcolor{teal}{+17.6} \\
Llama-3 (8B) & \textbf{98.4}\,{\tiny $\pm$ 0.6} & 97.9\,{\tiny $\pm$ 0.1} & \textcolor{brown}{-0.5} & 82.6\,{\tiny $\pm$ 1.9} & \textbf{90.1}\,{\tiny $\pm$ 0.6} & \textcolor{teal}{+7.5} & 48.9\,{\tiny $\pm$ 3.2} & \textbf{70.4}\,{\tiny $\pm$ 0.8} & \textcolor{teal}{+21.5} \\
Qwen-2.5 (14B) & 97.7\,{\tiny $\pm$ 0.3} & \textbf{99.2}\,{\tiny $\pm$ 0.1} & \textcolor{teal}{+1.5} & 86.2\,{\tiny $\pm$ 0.7} & \textbf{92.6}\,{\tiny $\pm$ 0.4} & \textcolor{teal}{+6.4} & 58.7\,{\tiny $\pm$ 1.5} & \textbf{73.0}\,{\tiny $\pm$ 0.8} & \textcolor{teal}{+14.3} \\
\bottomrule
\end{tabular}%
}
\vspace{-5pt}
\end{table*}
The results in Table~\ref{tab:model_scaling} indicate that raw parameter scaling alone does not resolve the deep-nesting bottleneck. While fine-tuning improves both methods, HLT yields markedly larger gains on deeper logic. Notably, an 8B model with HLT (70.4\%) significantly outperforms a larger 14B model using flat generation (58.7\%) on \texttt{D4\_B3}, highlighting structural decomposition as a more efficient lever for performance than scaling model capacity.

\subsection{Linguistic Diversity and Augmentation Scaling}

To analyze the effect of linguistic diversity, we fine-tune Llama-3-8B under different augmentation factors and compare Flat versus HLT across benchmark complexities.

\begin{table}[htpb]
\centering
\caption{Llama-3-8B Translation Accuracy vs. Linguistic Augmentation}
\label{tab:aug_ablation}
\resizebox{\columnwidth}{!}{%
\begin{tabular}{cl cccc}
\toprule
\multirow{2}{*}{\textbf{Dataset}} & \multirow{2}{*}{\textbf{Metric}} & \multicolumn{4}{c}{\textbf{Augmentation Factor}} \\
\cmidrule(lr){3-6}
& & \textbf{Aug = 1} & \textbf{Aug = 3} & \textbf{Aug = 5} & \textbf{Aug = 15} \\
\midrule
\multirow{3}{*}{\textbf{\texttt{D2\_B4}}} 
& \textbf{Flat (\%)} & 96.6\,{\tiny $\pm$ 0.4} & \textbf{97.6}\,{\tiny $\pm$ 1.0} & \textbf{98.4}\,{\tiny $\pm$ 0.6} & 97.8\,{\tiny $\pm$ 1.2} \\
& \textbf{HLT (\%)}  & 96.6\,{\tiny $\pm$ 1.1} & 96.9\,{\tiny $\pm$ 1.3} & 97.9\,{\tiny $\pm$ 0.1} & 97.8\,{\tiny $\pm$ 0.6} \\
& \multicolumn{1}{c}{\textbf{$\Delta$}}  & \textcolor{teal}{-0.0} & \textcolor{brown}{-0.7} & \textcolor{brown}{-0.5} & \textcolor{teal}{0.0} \\
\midrule
\multirow{3}{*}{\textbf{\texttt{D3\_B3}}} 
& \textbf{Flat (\%)} & 65.8\,{\tiny $\pm$ 3.2} & 76.6\,{\tiny $\pm$ 1.1} & 82.6\,{\tiny $\pm$ 1.9} & 83.3\,{\tiny $\pm$ 2.6} \\
& \textbf{HLT (\%)}  & \textbf{81.5}\,{\tiny $\pm$ 1.1} & \textbf{88.5}\,{\tiny $\pm$ 1.5} & \textbf{90.1}\,{\tiny $\pm$ 0.6} & \textbf{92.1}\,{\tiny $\pm$ 0.7} \\
& \multicolumn{1}{c}{\textbf{$\Delta$}}  & \textcolor{teal}{+15.7} & \textcolor{teal}{+11.9} & \textcolor{teal}{+7.5} & \textcolor{teal}{+8.8} \\
\midrule
\multirow{3}{*}{\textbf{\texttt{D4\_B3}}} 
& \textbf{Flat (\%)} & 34.0\,{\tiny $\pm$ 2.0} & 44.7\,{\tiny $\pm$ 1.5} & 48.9\,{\tiny $\pm$ 3.2} & 52.7\,{\tiny $\pm$ 1.4} \\
& \textbf{HLT (\%)}  & \textbf{50.4}\,{\tiny $\pm$ 1.2} & \textbf{64.0}\,{\tiny $\pm$ 1.1} & \textbf{70.4}\,{\tiny $\pm$ 0.8} & \textbf{68.2}\,{\tiny $\pm$ 4.6} \\
& \multicolumn{1}{c}{\textbf{$\Delta$}}  & \textcolor{teal}{+16.4} & \textcolor{teal}{+19.3} & \textcolor{teal}{+21.5} & \textcolor{teal}{+15.5} \\
\bottomrule
\end{tabular}%
}
\vspace{-5pt}
\end{table}

The scaling behaviors in Table~\ref{tab:aug_ablation} reveal that Flat generation improves linearly with data, yet consistently fails to bridge the gap with HLT in complex domains. Conversely, HLT converges rapidly, reaching near-peak performance (70.4\%) with a moderate data budget (\texttt{Aug=5}, 1,000 samples) and saturating thereafter. This early saturation confirms that structural priors mitigate the need for brute-force data scaling.

\subsection{Ablation: Data Allocation and Sample Efficiency}

To understand the source of HLT's performance gains and its practical data requirements, we conduct ablations on data allocation.

\textbf{Allocating Data Budgets.}
We compare maximizing \emph{formula diversity} (1,000 unique structures $\times$ 1 paraphrase) versus emphasizing \emph{linguistic diversity} (200 unique structures $\times$ 5 paraphrases, total 1,000 training pairs). For Flat generation, maximizing formula diversity is strictly necessary to improve on deep tasks (accuracy increases from 48.9\% to 59.3\% on \texttt{D4\_B3}). In stark contrast, HLT achieves superior performance (70.4\% vs. 70.0\%) using far fewer unique skeletons (200 vs. 1,000) exposed to rich paraphrasing. 

\textbf{Sample Efficiency.}
Furthermore, comparing Flat trained with 3,000 pairs against HLT with 600 pairs on \texttt{D4\_B3}, HLT achieves higher accuracy (64.0\% vs. 52.7\%) using \textbf{5$\times$ less data}. This confirms that HLT masters spatial logic with minimal structural examples, freeing up the model's capacity to absorb language variations.

\subsection{Downstream Robotic Applications}
\label{subsec:robotics_app}
To demonstrate practical utility beyond exact-match translation, we evaluate generated SpaTiaL formulas in downstream robotic pipelines for monitoring and action selection.

\textbf{Continuous Task Evaluation for VLA Executions.} 
We deploy generated SpaTiaL formulas to monitor multi-stage manipulation tasks within the LIBERO benchmark, executed by the $\pi_{0.5}$\cite{pi0.5} VLA policy. The explicit spatial predicates map to continuous kinematic observations (e.g., end-effector and object poses), producing a robustness signal $\rho(t)$ that quantifies satisfaction margins over time. As shown in Fig.~\ref{fig:stl_evaluation_all}, this continuous signal provides an interpretable evaluation of the VLA's performance. It explicitly identifies stage-wise progress in successful rollouts and quantifies geometric violations---such as repeated, unsuccessful grasping attempts---in failed executions, all without requiring manually designed reward functions.
\begin{figure}[ht]
    \centering
    \includegraphics[width=\columnwidth]{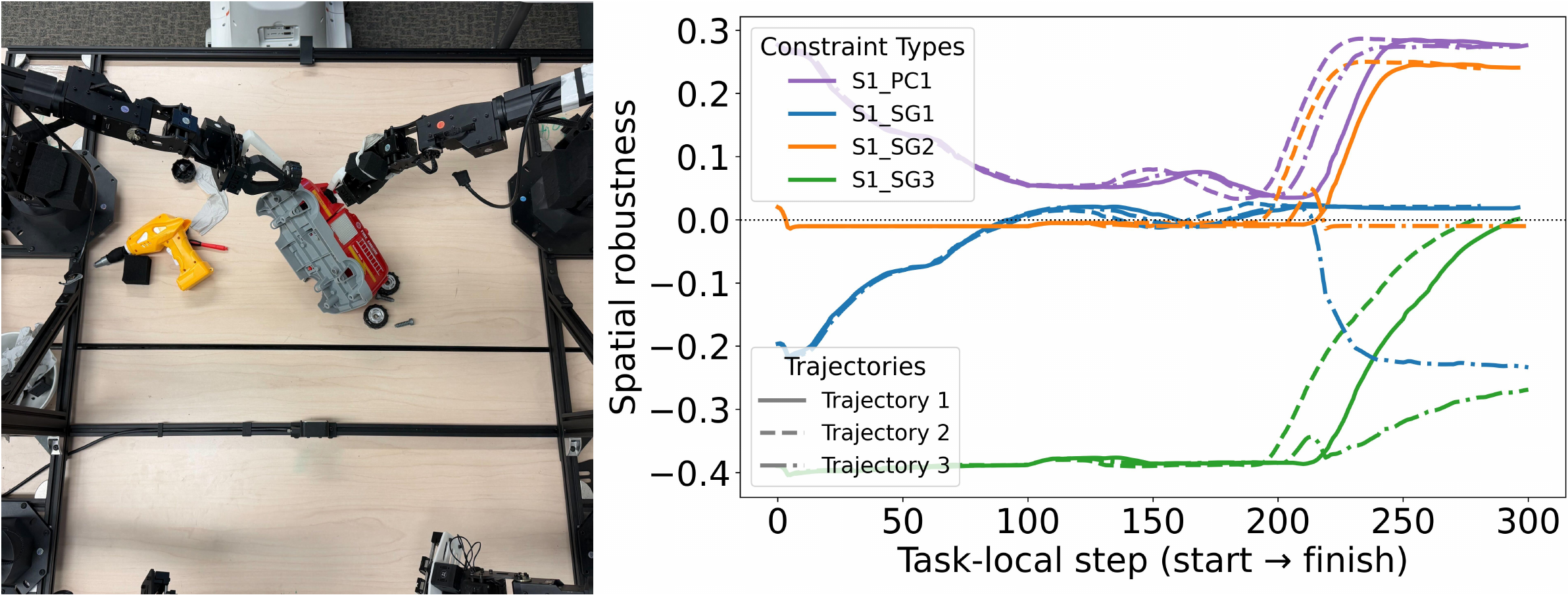}
    \caption{\textbf{Runtime verification of candidate VLA trajectories.} Short-horizon rollouts proposed by the $\pi_0$ model are explicitly scored against the HLT-translated SpaTiaL specification. }
    \vspace{-1.5em}
    \label{fig:Best_robustness_curve}
\end{figure}
\textbf{Runtime Trajectory Verification and Filtering.} 
Beyond passive monitoring, we evaluate the generated SpaTiaL formulas within a runtime ``generate-and-test'' pipeline using a VLA policy $\pi_0$~\cite{black2024pi0}. At each inference step, the unconstrained VLA proposes multiple candidate trajectories (short-horizon rollouts). We use the translated SpaTiaL specifications as explicit verifiers to score these rollouts. By rejecting trajectories that violate spatial constraints and selecting the one with the highest robustness (Fig.~\ref{fig:Best_robustness_curve}), the system effectively filters out dynamically unstable or geometrically unsafe future states. This confirms that our one-shot HLT translations successfully serve as executable, interpretable trajectory verifiers for reliable downstream deployment.

\FloatBarrier

\section{Conclusion}

We presented NL2SpaTiaL, a hierarchical framework for translating natural language into geometric Spatio-Temporal Logic specifications for manipulation tasks. By generating a structured logical tree followed by deterministic rendering, the method decouples semantic parsing from syntactic realization, improving robustness and sample efficiency over flat sequence generation, particularly for deeply nested logic. We also introduced a logic-first synthesis pipeline and a hierarchically annotated dataset with controllable complexity and diverse paraphrases, enabling systematic evaluation of structural reasoning in NL-to-logic translation.
A current limitation is that training primarily supervises the final nested JSON output, while richer hierarchical annotations—such as node-level descriptions and aligned intermediate subformulas—remain underutilized. Future work will incorporate multi-level supervision (e.g., node-level objectives or staged training), evaluate larger models under the same protocol, and extend the logic-first pipeline to pure TL benchmarks to test generality and scalability.

\bibliographystyle{IEEEtran}
\bibliography{ref}
\appendix

\subsection{Deterministic Rendering Policy}
\label{appendix:templates}

Table~\ref{tab:unified-templates} specifies the deterministic realization map $\tau(\cdot)$ used in our logic-first synthesis pipeline to convert SpaTiaL formulas into controlled English. Each template is deterministic, compositional, and preserves the information needed for round-trip supervision generation.

At the spatial level, templates expose geometric constraints explicitly (e.g., margins and tolerances). For temporal operators, the mapping preserves interval bounds verbatim. This deterministic design is critical for (i) generating semantically consistent synthetic training data and (ii) providing stable supervision targets for hierarchical NL-to-SpaTiaL translation.

\begin{table}[ht]
\centering
\caption{Unified deterministic templates}
\label{tab:unified-templates}
\resizebox{\columnwidth}{!}{%
\begin{tabular}{@{}ll@{}}
\toprule
\textbf{Operator} & \textbf{Deterministic English Template} \\
\midrule
\multicolumn{2}{@{}l}{\textit{Spatial Constraints}} \\
\midrule
$\Touch(i,j)$ & ``\ent{$i$} is in contact with \ent{$j$}.'' \\
$\CloseE(i,j)$ & ``The distance between \ent{$i$} and \ent{$j$} is at most $\varepsilon_c$.'' \\
$\FarE(i,j)$ & ``The distance between \ent{$i$} and \ent{$j$} is at least $\varepsilon_f$.'' \\
$\Ovlp(i,j)$ & ``\ent{$i$} partially overlaps \ent{$j$}.'' \\
$\PartOvlp(i,j)$ & ``\ent{$i$} overlaps \ent{$j$} without containment.'' \\
$\EnclIn(i,j)$ & ``\ent{$i$} lies strictly inside \ent{$j$}.'' \\
$\LeftOf(i,j)$ & ``\ent{$i$} is strictly to the left of \ent{$j$} (margin $\kappa$).'' \\
$\Between_{\mathrm{px}}(a,b,c)$ & ``Along the $x$-axis, \ent{$b$} lies strictly between \ent{$a$} and \ent{$c$}.'' \\
$\text{oriented}(i,j;\kappa)$ & ``The heading of \ent{$i$} is aligned with that of \ent{$j$} (within $\kappa$).'' \\
\midrule
\multicolumn{2}{@{}l}{\textit{Temporal / Boolean Logic}} \\
\midrule
$G_{[a,b]} \,\phi$ & ``Throughout \ival{a}{b}, $\tau(\phi)$ holds.'' \\
$F_{[a,b]} \,\phi$ & ``Sometime within \ival{a}{b}, $\tau(\phi)$ holds.'' \\
$\phi_1 \,U_{[a,b]} \,\phi_2$ & ``Within \ival{a}{b} $\tau(\phi_2)$ becomes true; until then, $\tau(\phi_1)$ holds.'' \\
$\neg\,\phi$ & ``It is not the case that $\tau(\phi)$.'' \\
$\phi_1 \wedge \phi_2$ & ``$\tau(\phi_1)$ and $\tau(\phi_2)$ both hold.'' \\
$\phi_1 \vee \phi_2$ & ``Either $\tau(\phi_1)$ or $\tau(\phi_2)$ holds.'' \\
$\phi_1 \rightarrow \phi_2$ & ``If $\tau(\phi_1)$ holds, then $\tau(\phi_2)$ must hold.'' \\
\bottomrule
\end{tabular}%
}
\vspace{-15pt}
\end{table}







\end{document}